\documentclass[lettersize,journal]{IEEEtran}
\usepackage{epsfig,rotating,setspace,latexsym,amsmath,epsf,amssymb,amsfonts,bm,theorem,booktabs,subfigure,epstopdf}

\usepackage{cite,authblk,tikz}
\usepackage{color}

\definecolor{commentcolor}{RGB}{110,154,155}   % define comment color
\IEEEoverridecommandlockouts
\allowdisplaybreaks

\begin{document}
\bstctlcite{IEEEexample:BSTcontrol} 
\title{Continual Deep Reinforcement Learning to Prevent Catastrophic Forgetting in Jamming Mitigation} 
\author{Kemal Davaslioglu}
\author{Sastry Kompella}
\author{Tugba Erpek}
\author{Yalin E. Sagduyu}
%\author[2]{Ananthram Swami}
\affil{\normalsize Nexcepta, Gaithersburg, MD, USA 
\thanks{This material is based upon work supported by the ASA(ALT) SBIR CCoE under Contract No. W51701-23-C-0145.}
} 
\maketitle

\begin{abstract}
	Deep Reinforcement Learning (DRL) has been highly effective in learning from and adapting to RF environments and thus detecting and mitigating jamming effects to facilitate reliable wireless communications. However, traditional DRL methods are susceptible to catastrophic forgetting (namely forgetting old tasks when learning new ones), especially in dynamic wireless environments where jammer patterns change over time. This paper considers an anti-jamming system and addresses the challenge of catastrophic forgetting in DRL applied to jammer detection and mitigation. First, we demonstrate the impact of catastrophic forgetting in DRL when applied to jammer detection and mitigation tasks, where the network forgets previously learned jammer patterns while adapting to new ones. This catastrophic interference undermines the effectiveness of the system, particularly in scenarios where the environment is non-stationary. We present a method that enables the network to retain knowledge of old jammer patterns while learning to handle new ones. Our approach substantially reduces catastrophic forgetting, allowing the anti-jamming system to learn new tasks without compromising its ability to perform previously learned tasks effectively. Furthermore, we introduce a systematic methodology for sequentially learning tasks in the anti-jamming framework. By leveraging continual DRL techniques based on PackNet, we achieve superior anti-jamming performance compared to standard DRL methods. Our proposed approach not only addresses catastrophic forgetting but also enhances the adaptability and robustness of the system in dynamic jamming environments. We demonstrate the efficacy of our method in preserving knowledge of past jammer patterns, learning new tasks efficiently, and achieving superior anti-jamming performance compared to traditional DRL approaches.  
\end{abstract}


\section{Introduction}\label{sec:Introduction}
The proliferation of wireless communication systems has been accompanied by an escalating threat landscape, particularly in the form of jamming attacks. Due to the open and shared nature of wireless medium, jamming, an intentional interference with wireless signals, poses a significant threat to the reliability, efficiency, and security of wireless networks. It disrupts the normal functioning of communication systems, leading to degraded service quality or complete denial of service. The dynamic and complex nature of wireless channels, coupled with the evolving sophistication of jamming techniques, necessitates advanced solutions for jammer detection and mitigation \cite{Sagduyu2008, pirayesh2022jamming, costa2023timely}. 

Machine learning (ML) offers a promising avenue for addressing these challenges due to its ability to learn and model complex patterns and behaviors. Specifically, the application of ML for anti-jamming allows for an adaptive and intelligent approach to counteract jamming effects. This adaptability is crucial in dealing with the variability of channel conditions and the characteristics of jammers, which traditional rule-based systems may fail to effectively counter \cite{davaslioglu2019deepwifi, erpek2018deep, Shi2018, pourranjbar2021reinforcement}. 

Among various ML techniques, reinforcement learning (RL) stands out for its potential in anti-jamming. RL differs from other machine learning approaches by learning optimal behaviors through interactions with the environment, rather than relying on a pre-labeled dataset. This feature is particularly advantageous for anti-jamming strategies, where the environment (i.e., the wireless channel and the jammer's behavior) is dynamic and unpredictable. RL enables the system to learn and adapt to jamming effects on the fly, offering a robust solution for real-time detection and mitigation of jammers, even in the absence of supervised training data that may not be available for zero-day jammer threats \cite{yao2019collaborative, yang2020intelligent, pourranjbar2021reinforcement}. 

Deep reinforcement learning (DRL), combining deep learning and RL, further enhances the capability to deal with the complexities of anti-jamming \cite{han2017two, liu2018anti, abuzainab2019qos, wang2021jamming}. By leveraging deep neural networks, DRL can model highly complex strategies and patterns, outperforming traditional RL in environments with vast state and action spaces such as those encountered in wireless communication systems. This makes DRL an ideal candidate for developing sophisticated anti-jamming strategies that can adapt to a wide range of jamming scenarios. 

However, the application of DRL in dynamic environments like wireless communications is not without challenges. A significant issue is catastrophic forgetting (catastrophic interference), a phenomenon where ML models and RL mechanisms forget previously learned tasks upon learning new tasks \cite{kirkpatrick2017overcoming}. Catastrophic forgetting has been demonstrated for signal classification in wireless systems \cite{shi2019deep}, where a receiver may learn to classify new types of signals (such as new modulations) while forgetting old ones using elastic weight consolidation (EWC) \cite{kirkpatrick2017overcoming}. Continual learning with EWC has been also applied to wireless systems in the context of synchronization in digital twins over wireless networks \cite{omar2022edge}. Catastrophic forgetting is particularly problematic in anti-jamming applications, where the ability to remember previously encountered jammer strategies and adapt to new jammer patterns is crucial for sustained system performance.  

To address this critical challenge, we propose the use of progressive neural networks such as PackNet as a solution to prevent catastrophic forgetting in DRL-based anti-jamming systems. PackNet introduces a structured approach to neural network parameter management, allowing the network to retain knowledge of old tasks while learning new ones efficiently \cite{mallya2018packnet}. By employing PackNet in the context of DRL for anti-jamming, we aim to develop a robust, adaptive, and continuously learning system capable of mitigating the effects of a wide range of jamming attacks without forgetting previously learned strategies. This paper presents our continual DRL solution with PackNet for anti-jamming, focusing on the challenges of catastrophic forgetting and the strategies to overcome them in the dynamic and adversarial environment of wireless communications.

The novel contributions of this paper are given as follows:

\begin{enumerate}

	\item \textit{Enhanced System Adaptability and Robustness:} Our approach, which combines continual deep reinforcement learning (DRL) techniques with PackNet, significantly improves the system's capacity to adjust and manage various jamming scenarios as they occur. By fostering a more flexible and resilient framework, our methodology bolsters the robustness of wireless networks, ensuring they remain effective in the face of diverse and changing jamming challenges.

	\item \textit{Effective Mitigation of Catastrophic Forgetting:} A core contribution of our work is addressing the challenge of catastrophic forgetting, which is prevalent in standard DRL applications. Our proposed framework successfully retains critical knowledge of previously encountered jamming patterns while seamlessly incorporating new information. This capability ensures sustained system performance over time, without the degradation typically associated with learning new tasks in neural network models.

	\item \textit{Superior Anti-Jamming Performance:} Through comprehensive experiments and simulations, we demonstrate that our approach not only preserves knowledge effectively but also achieves a higher level of anti-jamming performance compared to traditional DRL methods. This is quantitatively evidenced by improved metrics in detecting and mitigating a wide range of jamming attacks, illustrating the practical benefits of our methodology in enhancing the security and reliability of wireless networks. 
	
	\item \textit{Reward Engineering:} As RL-based AI systems become more autonomous, the design of appropriate reward mechanisms that elicit desired behaviors becomes more critical. Reward engineering \cite{dewey2014reweng} is increasingly becoming an important part of the DRL framework. For the anti-jamming framework considered in this paper, we assign the scaled and normalized spectral efficiency as the reward function. This definition provides a physics-informed reward function that incentives the AI agent to efficiently utilize the available communication resources while mitigating jamming effects.
	
	\item \textit{Efficiency in Learning New Tasks:} Our research underscores the efficiency of our proposed method in learning new jammer patterns and strategies. The integration of PackNet allows for a structured and efficient approach to task learning, minimizing the resources required for adaptation while maximizing the retention of valuable knowledge. This efficiency is crucial for the practical deployment of anti-jamming technologies in real-world scenarios, where computational resources and response times are often limited.

\end{enumerate}

These contributions represent a significant advancement in the field of anti-jamming systems, offering a robust and adaptable solution to the ever-evolving threat of jamming. Our work not only addresses the theoretical challenges associated with DRL in the context of catastrophic forgetting but also provides a practical framework for the development of next-generation anti-jamming systems. 

The remainder of this paper is organized as follows. Section~\ref{sec:system_model} presents system model. Section~\ref{sec:scenarios} elaborates on the scenarios considered and developed in this paper. Section~\ref{sec:continual_learning} presents the proposed continual learning framework. Section~\ref{sec:performance} describes the performance evaluations and discusses the obtained results. Section~\ref{sec:Conclusion} concludes the paper. 

\section{System Model}\label{sec:system_model}
We provide a succinct overview of the DRL agents utilized in this paper, namely, the Deep Q-network (DQN) \cite{mnih2013atari} and Soft Actor-Critic (SAC) \cite{haarnoja2018sac} algorithms. The DQN combines Q-learning with deep neural networks to approximate and optimize the Q-values. DQN enables agents to make decisions in environments with high-dimensional state spaces. The goal of DRL agent is to interact with the environment and select actions in a way that maximizes its future rewards. The future rewards of the agent are discounted by a factor of $\gamma$ per time-stamp where $\gamma\in (0,1]$ and the future discounted return at time $t$, $R_{t_0}$, is defined as 
\begin{align}
	R_{t_0} = \sum_{t'=t_{0}}^\infty \gamma^{t-t_0} r_{t},
\end{align}
where $r_t$ is the reward at time $t$. The optimal action-value function, $Q^*(s,a)$ can be defined as the maximum expected return achievable by following any strategy $\pi$ after seeing some sequence of states $s$ and taking some actions $a$. The stochastic policy $\pi$ learns to map sequences to actions or distributions over actions, which can be represented as 
% \matchcal{S} \times \mathcal{A} \rightarrow \mathbb{R}
\begin{align}
	Q^*(s,a) = \max_\pi \mathbb{E} \left[ R_t | s_t=s, a_t=a, \pi \right],
\end{align}
where $s_t$ and $a_t$ are the state visited and the action taken at time $t$, respectively. The set of all possible states is shown by $\mathcal{S}$, and $a_t$ is selected from some set of possible actions $\mathcal{A}$.

SAC introduces a maximum entropy framework, promoting exploration and enabling better handling of stochastic environments by optimizing both the policy and the value function. The SAC framework extends the standard reinforcement learning objective of maximizing the return by simultaneously maximizing the entropy of the agent's policy, which can be expressed as
\begin{align}
	\pi^\star = \arg \max_\pi \mathbb{E}_\pi \left[ \sum_{t=0}^{\infty} \gamma^t \left(r(s_t,a_t) + \alpha \mathcal{H} \left( \pi \left( \cdot | s_t \right) \right) \right)\right],
\end{align}
where the reward at time $t$ is expressed by $r(s_t, a_t)$. The entropy is denoted by 
$\mathcal{H}$ and $\alpha$ is used to control the balance between reward and entropy maximization. SAC formulation has a close connection to the exploration-exploitation trade-off \cite{sutton2018reinforcement} where  increasing entropy results in more exploration, but it accelerates learning later  and prevents the policy from prematurely converging to a bad local optimum. 

\subsection{Scenarios}\label{sec:scenarios}
For jamming detection and mitigation, we define different scenarios, each presenting increasing levels of difficulty.

\begin{table}[tb!]
	\centering
	\caption{Jamming pattern in Env.~1.}
	\label{table:sweep_jamming}
	\footnotesize
	\begin{tabular}{c|cccc}
		\toprule 
		Time & Ch~1 & Ch~2 & Ch~3 & Ch~4 \\ \midrule 
		$t$ & 1.0 & 0.0  & 0.0 & 0.0 \\ 
		$t+1$ & 0.0 & 1.0 & 0.0 & 0.0 \\ 
		$t+2$ & 0.0 & 0.0 & 1.0 & 0.0 \\ 
		$t+3$ & 0.0 & 0.0 & 0.0 & 1.0 \\ 				
		\bottomrule
	\end{tabular}
\end{table}

\textbf{Scenario 1:} In the first scenario, we define Environment 1 (Env~1) that considers a linear sweeping jamming pattern where the jammer selects one of $N$ channels and transmits with full power only on this channel. In the next time instant, the jammer switches to the next channel and again transmits at full power only in this channel. This process is repeated until all $N$ channels are jammed in $N$ time instants. This jamming pattern in Env~1 is shown in Table~\ref{table:sweep_jamming}. The goal of the agent is to learn the jamming pattern and select a next time slot that is interference-free to mitigate the adverse effects of interference. For this scenario, we identify the following states, actions, rewards, and game over conditions:
\begin{itemize}
	\item States: The channel occupancy of all $N$ channels in the current time slot, i.e., wideband sensing.
	\item Action: Agent selects the channel to be used for the next time slot. 
	\item Reward: The agent receives a $+1$ reward if there is no collision with interferer; otherwise, $-1$ penalty is incurred. 
	\item Game over condition: Each epoch runs until the agent makes three collisions in the last ten time slots or maximum number of steps is reached. 
\end{itemize}

\textbf{Scenario 2:} In this scenario, the jamming behavior changes from the one in Env~1 to a new jamming pattern where the jammer uses a non-uniform power allocation across different number of channels that are selected randomly. We refer to this environment as Env~2. Table~\ref{table:jamming_pattern2} presents the jamming pattern considered in Env~2. In this scenario, as the jammer can interfere with multiple channels, we need to update the action, reward, and game over condition definitions. For the actions, the agent can select a channel to transmit and a Modulation Coding Scheme (MCS) to adapt its waveform. Also, we update the reward definition from a simple $\pm 1$ to a more elaborate spectral efficiency, that is typically represented in bits/sec/Hz. As reinforcement learning environments typically use positive rewards to encourage behavior and negative ones for the negative feedback, we apply standard normalization the spectral efficiency of the data to obtain the reward function. This also enables us to introduce a \emph{physics-inspired} reward based to adjust the waveform based on the predicted SINR and incorporate the effects of error correction and resiliency. For this scenario, the state, action, reward, and game over conditions are defined as follows:

\begin{enumerate}
	\item State: The channel occupancy of all $N$ channels in the current time slot. In this scenario, jammer uses a more complex jamming pattern that is shown in Table~\ref{table:jamming_pattern2}.
	\item Action: The agent selects a channel to transmit and an MCS value to optimize the data transmission. 
	\item Reward: The jammer can interfere with the transmissions at different power levels, we need to update the simple reward definition in Scenario~1 to a normalized and scaled spectral efficiency, that is shown in Table~\ref{table:reward_se_sinr}. 
	\item Game over condition: In this environment, there are time instances where the agent cannot find an empty channel due to the interference caused by the jammer. To encourage the agent to select channels with less interference, we need to define collision. For this scenario, we consider a collision if the inference level on the channel exceeds 0.25. Similar to Env~1, if we observe three collisions in the past ten time slots or the maximum number of steps is reached, the game is terminated. 		
\end{enumerate}

\begin{table}[tb!]
	\centering
	\caption{Jamming pattern in Env~2.}
	\label{table:jamming_pattern2}
	\footnotesize
	\begin{tabular}{c|cccc}
		\toprule 
		Time & Ch~1 & Ch~2 & Ch~3 & Ch~4 \\ \midrule 
		$t$ & 0.500 & 0.125 & 0.250 & 0.000 \\ 
		$t+1$ & 0.250 & 0.000 & 0.500 & 0.375 \\ 
		$t+2$ & 0.125 & 0.375 & 0.000 & 0.500 \\ 
		$t+3$ & 0.375 & 0.000 & 0.500 & 0.250 \\ 		
		$t+4$ & 0.125 & 0.375 & 0.500 & 0.250 \\ 
		$t+5$ & 0.250 & 0.500 & 0.125 & 0.000 \\ 
		$t+6$ & 0.250 & 0.375 & 0.500 & 0.000 \\ 
		$t+7$ & 0.375 & 0.250 & 0.125 & 0.500 \\ 
		$t+8$ & 0.000 & 0.125 & 0.375 & 0.500 \\ 
		$t+9$ & 0.000 & 0.250 & 0.125 & 0.375 \\ 				
		\bottomrule
	\end{tabular}
\end{table}

\begin{table}[tb!]
	\centering
	\caption{Reward values and spectral efficiency (SE) as a function of channel and SINR.}
	\footnotesize
	\label{table:reward_se_sinr}
	\begin{tabular}{c|cccc}
		\toprule 
		Interference & MCS & SINR & SE & Reward \\
		& & (dB) & (b/s/Hz) & \\ \midrule
		0.000 & BPSK + 1/2 & 12 & 0.5 & 0.1 \\
		0.000 & QPSK + 1/2 & 12 & 1.0 & 1.0 \\
		0.000 & 16QAM + 1/2 & 12 & 1.5 & 2.0 \\
		
		0.125 & BPSK + 1/2 & 10 & 0.5 & 0.1 \\
		0.125 & QPSK + 1/2 & 10 & 1.0 & 0.6 \\
		0.125 & 16QAM + 1/2 & 10 & 1.5 & 1.0 \\
		
		0.250 & BPSK + 1/2 & 7 & 0.5 & 0.1 \\
		0.250 & QPSK + 1/2 & 7 & 0.7 & 0.4 \\
		0.250 & 16QAM + 1/2 & 7 & 1.0 & 0.2 \\
		
		0.375 & BPSK + 1/2 & 6 & 0.3 & -0.4 \\
		0.375 & QPSK + 1/2 & 6 & 0.6 & -0.1 \\
		0.375 & 16QAM + 1/2 & 6 & 0.7 & -0.2 \\
		
		0.500 & BPSK + 1/2 & 4 & 0.2 & -0.8 \\
		0.500 & QPSK + 1/2 & 4 & 0.3 & -1.0 \\
		0.500 & 16QAM + 1/2 & 4 & 0.0 & -1.0 \\
		
		1.000 & BPSK + 1/2 & 0 & 0.0 & -1.6 \\
		1.000 & QPSK + 1/2 & 0 & 0.0 & -1.8 \\
		1.000 & 16QAM + 1/2 & 0 & 0.0 & -2.0 \\
		\bottomrule
	\end{tabular}
\end{table}

\textbf{Section~3A:} 
Scenario~3A is an extension of Scenario~2 where the DRL agent initially learns Env~1 and then Env~2. Jammer changes its behavior back to the one in Env~1. The question that we seek to answer is \emph{Does the agent adapt back to remembering the initial task it has learned after adapting to learn the second task?}	 This condition helps us validate if there is any catastrophic forgetting case or not. We present the state, action, reward, and game over condition definitions for completeness:
\begin{enumerate}
	\item State: Current channel sensing results as in Scenario~2.
	\item Actions: The agent selects a channel to transmit and the MCS to adjust its waveform.
	\item Reward: The agent maximizes the normalized and scaled spectral efficiency based on SINR in Table~\ref{table:reward_se_sinr}.  
	\item Game over condition: It is the same as in Env~2. 
\end{enumerate}

\textbf{Scenario 3B:} Similar to Scenario~3A, Scenario~3B is an extension of Scenario 2 where the DRL agent initially learns Env 1 and then Env 2. In Scenario~3B, jammer changes its behavior to a new environment that we denote as Env~3, merging its jamming behavior in Env~1 and Env~2. We can express the state, action, reward, and game over condition definitions as follows:
	\begin{enumerate}
		\item State: Current channel sensing results as in Scenario~2. 
		\item Action: The agent selects a channel to transmit and the MCS to adjust its waveform.
		\item Reward: The agent maximizes the normalized and scaled spectral efficiency based on SINR in Table~\ref{table:reward_se_sinr}.
		\item Game over condition: It is the same as in Env~2. 
	\end{enumerate}

\section{Continual Learning Framework}\label{sec:continual_learning}

In this section, we describe the parameter isolation approach that we employed for anti-jamming. This approach involves several key steps:  

First, for task $n$, $T_n$, we train the network while keeping the parameters $\theta_{1:n-1}$ associated with previous tasks frozen. Following the training phase, we proceed to pruning, wherein a fraction of the network weights are set to zero. This is accomplished by assigning a binary mask to allocate a subset of parameters. Rather than randomly selecting weights for pruning, we sort the weights in each layer (convolutional and/or fully connected) based on their absolute values, indicating their importance. We then discard either the lowest 50\% or 75\%, following the suggested numbers in \cite{mallya2018packnet,DSD16}. 

To maintain simplicity, we adopt a one-shot pruning method, although incremental pruning has shown promise for improved results \cite{han2015learning}. Notably, extensive pruning, especially with high pruning ratios, can lead to an immediate performance drop due to significant changes in network connectivity \cite{mallya2018packnet}. To mitigate this, we perform a small number of re-training steps where we retrain the most crucial parameters $\theta_{n}$ while keeping the parameters of previous tasks $\theta_{1:n-1}$ masked out. This ensures that the performance on task $T_n$ is preserved during inference, utilizing only unmasked parameters $m_{1:n}$.

After a round of pruning and re-training, we achieve a network with sparse filters and minimal performance degradation on task $T_n$. Importantly, during the pruning step, we only remove weights associated with the current task, leaving those from previous tasks untouched. This ensures that introducing a new task does not compromise the performance of prior tasks.

\section{Performance Evaluation } \label{sec:performance}
In our numerical evaluations, we used the ns-3 gym environment. Ns-3 \cite{ns3} is an open-source discrete-event simulator for network systems. For many RL applications, OpenAI's Gym \cite{openai_gym} is used to support the development of RL agents for a variety of applications ranging from playing video games like Pong or Pinball to robotics applications. Due to the Gym’s easy interface, it is commonly used by different ML frameworks. The ns-3 simulator framework is integrated with the Gym using the ns3-gym environment \cite{ns3gym}. We have incorporated different jamming strategies by expanding the sweeping jammer code that was proposed in the ns3-gym library and making our adaptive jamming pattern module to develop jammers with different jamming behaviors. 
	
\subsection{Baseline Scenario}
\begin{figure}[tbh!]
	\centering
	\includegraphics[width=\columnwidth]{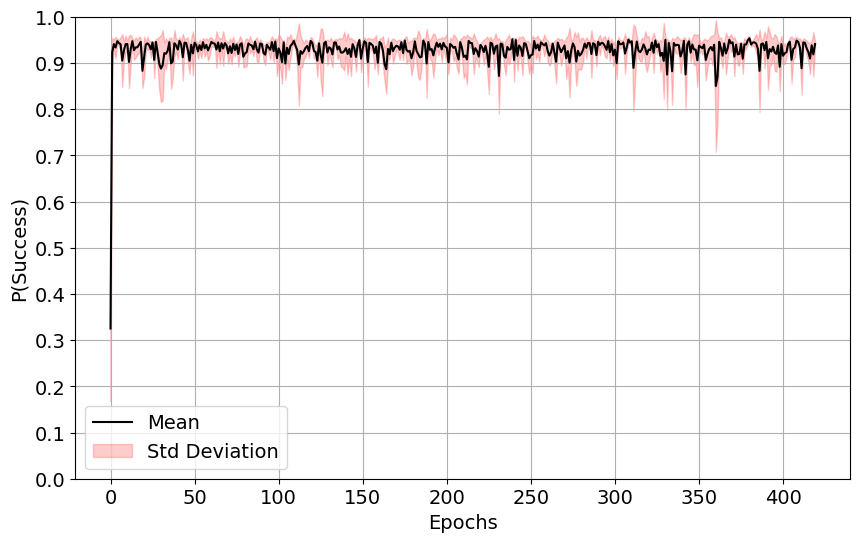}
	\caption{DQN performance in Scenario~1.}
	\label{fig:sc1_dqn}
\end{figure}

\begin{figure}[tbh!]
	\centering
	\begin{tikzpicture}
		\node (image1) at (0,0) {\includegraphics[width=\linewidth]{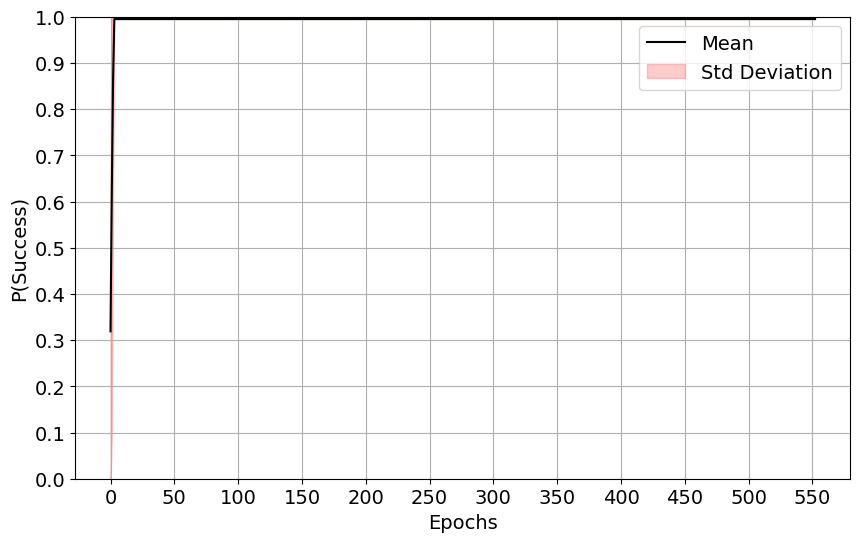}};
		\node (image2) at ([yshift=-0.05\linewidth, xshift=0.15\linewidth] image1.center) {\includegraphics[width=0.6\linewidth]{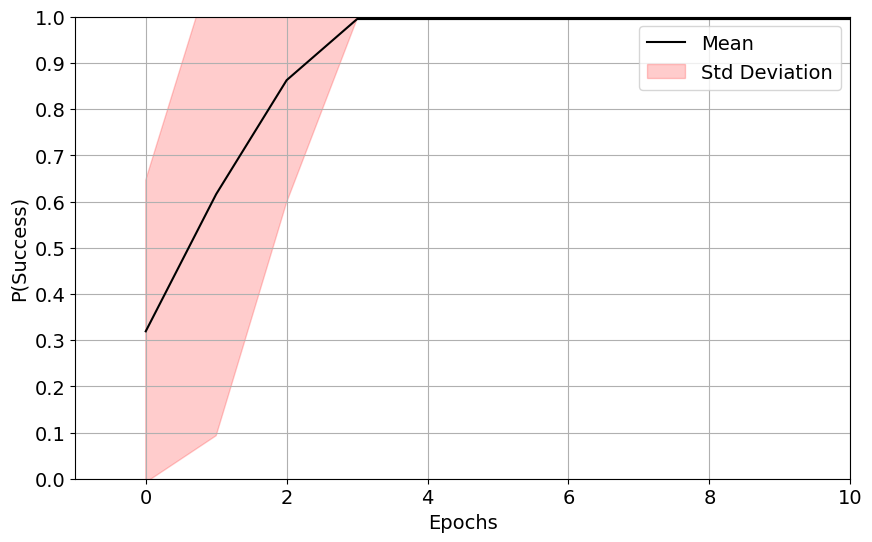}};
	\end{tikzpicture}
	\caption{SAC performance in Scenario~1.}
	\label{fig:sc1_sac}
\end{figure}

In Scenario~1, we train two DRL agents using the DQN and SAC approaches described in Section~\ref{sec:system_model}. We measure the average reward obtained by the agents over 450 epochs. We repeat this process five times with different seeds to ensure robustness and reliability of the results. Monitoring the average reward provides a general indication of the performance of the DRL agents. Figs.~\ref{fig:sc1_dqn} and \ref{fig:sc1_sac} demonstrate the mean reward of DQN and SAC agents at different epochs averaged over five seeds, respectively. These results have demonstrated that both agents in Scenario~1 are very successful, mostly due to the fact that the jamming pattern is very predictable. For example, the DQN agent quickly learns the jamming pattern and achieves more than 95\% success probability in selecting the channel without interference. SAC agent is able to achieve an average reward with a 100\% success rate which indicates it was able to select a channel without any collision over all five runs. Furthermore, it took around only three epochs for the SAC agent to fully learn the jamming pattern and avoid the interference. 

% Scenario 1

% Scenario 1

\subsection{Continual Learning Scenarios}
In the continual learning scenarios, we adopt the DQN agent due to its simplicity and similar performance. To measure the continual learning capabilities, we consider three different baseline initializations which are described as follows: 
\begin{enumerate}
	\item \textbf{No pretraining:} DRL agent resets its coefficients, empties its buffer, and starts to learn the new environment with new observations.
	\item \textbf{Pretrained:} DRL agent uses the weights and biases learned in previous task and updates the network with new observations in the new environment.
	\item \textbf{PackNet:} DRL agent uses the weights and biases learned in previous task, applies pruning, and finetunes the network with limited observations. 
\end{enumerate}
All three DRL agents have the exact same neural network architecture for a fair assessment, and they are trained using the same number of epochs, learning rates, and set of seeds for random number generation. We implemented a fully connected feedforward neural network with 6 layers, where the hidden layers have 256 neurons in each layer and Rectifying Linear Unit (ReLU) activations are used in between layers. The AdamW optimizer \cite{adam_optimizer} is used with a learning rate of 0.01. 

\subsubsection{Scenario 2}
\begin{figure}[tbh!]
	\centerline{\includegraphics[width=\linewidth,height=2in]{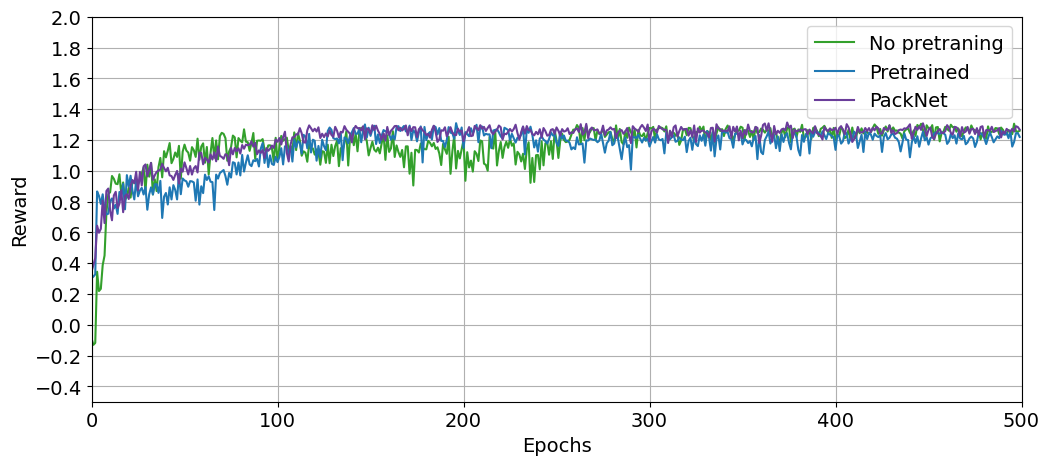}}
	\caption{Mean reward performance of three agents in Scenario~2.}
	\label{fig:sc2}
\end{figure}

In this scenario, No pretraining DRL agent starts fresh to learn the Env~2. Pretrained DRL agent uses the parameters learned in Env~1 and adapts to Env~2 by interacting with the new environment. PackNet DRL agent applies the continual learning framework described in Section~\ref{sec:continual_learning} with pruning. Fig.~\ref{fig:sc2} presents the average rewards over epochs and Table~\ref{table:sc2_reward} summarizes the mean reward results over five seeds. In this scenario, we can see that all agents perform similarly and rapidly learn the new jamming behavior in Env~2. PackNet DRL Agent slightly provides better performance in Env~2 compared to the other two agents. 
	
\begin{table}[tbh!]
 \footnotesize
	\centering
	\caption{Scenario 2: The first environment change (Env1 $\rightarrow$ Env2).}
	\label{table:sc2_reward}
	\begin{tabular}{cc}
		\toprule
		DRL Type & Mean reward over 5 seeds \\ \hline 
		No pretraining & 1.06 ($\pm$ 0.23) \\
		Pretrained & 1.07 ($\pm$ 0.26) \\
		PackNet & 1.12 ($\pm$ 0.20) \\
		\bottomrule
	\end{tabular}
\end{table}

\subsubsection{Scenario 3A}
In this scenario, the jammer in Scenario~2 changes its behavior back to the one in Env~1. For all three DRL agents, the network parameters at the end of Scenario~2 are frozen when we evaluate them in this scenario. Table~\ref{table:sceneario_3a} presents the mean reward over five seeds for the three DRL agents. Ideally, the Pretrained DQN agent has learned the jammer behavior in Env~1 and Env~2, so when the jammer switches back to Env~1, it should be able to remember the environment and perform well. However, we observe that its performance is as bad as the No pretraining DRL agent that has not seen the jammer behavior in Env~1, where both agents perform around a mean reward of 0.080. This highlights a \emph{catastrophic forgetting scenario} in the RL setting where the agent forgets the knowledge it has acquired in a previously trained task as it learns a new one. PackNet, on the other hand, provides the best resistance among the three baselines and maintains %the mean reward that achieves 
a mean reward of 0.829. 

% Scenario 3A
\begin{table}[htb!]
 \footnotesize
	\centering
	\caption{Scenario 3A: Environment changes from Env~2 back to Env~1 (Env~2 $\rightarrow$ Env~1).}
	\label{table:sceneario_3a}
	\begin{tabular}{cc}
		\toprule
		DRL Type & Mean reward over 5 seeds \\ \hline 
		No pretraining & 0.080 ($\pm$ 0.226) \\
		Pretrained & 0.079 ($\pm$ 0.451) \\
		PackNet & 0.829 ($\pm$ 0.597) \\
		\bottomrule
	\end{tabular}
\end{table}

\subsection{Scenario 3B}
\begin{figure}[tbh!]
	\centerline{\includegraphics[width=\linewidth,height=2in]{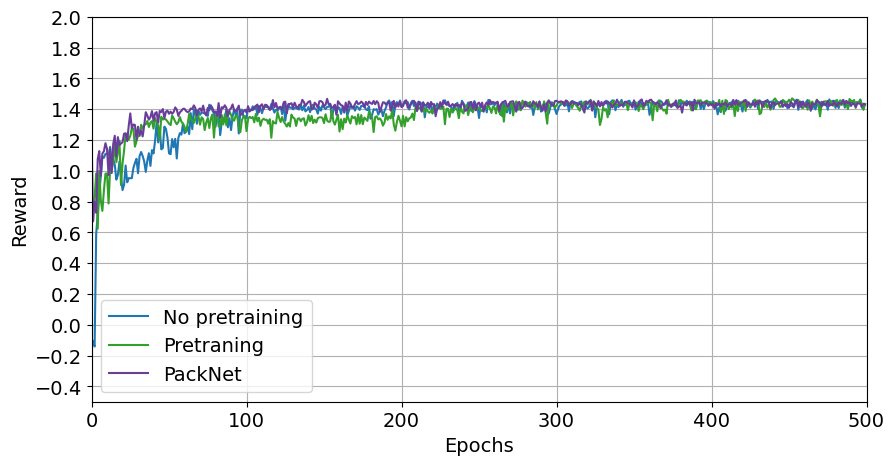}}
	\caption{Mean reward of the three DRL agents in Scenario~3B.}
	\label{fig:sc3b}
\end{figure}

In this scenario, the jammer adapts a new behavior, labeled as Env~3, where it integrates its jamming tactics from both Env~1 and Env~2. This scenario is designed to highlight the sequential task learning capabilities of the agents. Fig.~\ref{fig:sc3b} shows the mean reward of all three DRL agents over 500 epochs. We observe that all the agents achieve similar performance and converge to the same level given enough training epochs. However, the PackNet method approaches  this point 20-40 epochs faster than both methods. Table~\ref{table:scenario3b} presents the mean over the 500 epochs which shows that the PackNet method yields better overall operational performance due to its faster convergence. 

% Scenario 3B

\begin{table}[tbh!]
 \footnotesize
	\centering
	\caption{Scenario 3B: Environment changes from Env~2 to Env~3 \\ (Env~2 $\rightarrow$ Env~3).}\label{table:scenario3b}
	\begin{tabular}{cc}
		\toprule
		DRL Type & Mean reward over 5 seeds \\ \hline 
		No pretraining & 1.367 ($\pm$ 0.179) \\
		Pretrained & 1.365 ($\pm$ 0.124) \\
		PackNet & 1.405 ($\pm$ 0.101) \\
		\bottomrule
	\end{tabular}
\end{table}

\section{Conclusion} \label{sec:Conclusion}
In this paper, we study the challenge of catastrophic forgetting for anti-jamming systems, where DRL is applied to jammer detection and mitigation tasks within dynamic RF environments. Traditional DRL methods struggle with retaining knowledge of previously learned jammer patterns while adapting to new ones, thereby compromising system effectiveness, particularly in non-stationary environments. To address this issue, we studied the parameter isolation method to mitigate catastrophic forgetting. Our approach enables the network to preserve knowledge of old jammer patterns while effectively learning to handle new ones, thus significantly reducing interference and enhancing system performance. Additionally, we introduced a systematic approach for sequentially learning tasks in the anti-jamming framework, leveraging continual DRL techniques based on PackNet. We demonstrated the effectiveness of our approach in preserving past knowledge, efficiently learning new tasks, and achieving superior anti-jamming performance compared to traditional DRL approaches. This approach contributes to the advancement of anti-jamming systems by improving adaptability and robustness of wireless communication systems in dynamic jamming environments. 

\bibliographystyle{IEEEtran}
\bibliography{refs2}
	
\end{document}